\documentclass{bmvc2k}

\usepackage{amssymb}
\usepackage{amsmath}
\usepackage{pifont}
\usepackage{dirtytalk}
\usepackage{tabulary}
\usepackage{arydshln}
\usepackage{bm}

\usepackage{overpic}
\usepackage{multirow}
\usepackage{booktabs}

\newlength\savewidth

\newcommand{\tablestyle}[2]{\setlength{\tabcolsep}{#1}\renewcommand{\arraystretch}{#2}\centering\small}

\title{Feature and Label Embedding Spaces Matter in Addressing Image Classifier Bias}

\addauthor{William Thong \textsuperscript{$\ast$} }{w.e.thong@uva.nl}{1}
\addauthor{Cees G. M. Snoek}{cgmsnoek@uva.nl}{1}

\addinstitution{
 University of Amsterdam\\
 Science Park 904\\
 Amsterdam\\
 The Netherlands
}

\runninghead{Thong and Snoek}{Feature and Label Spaces Matter in Bias Mitigation}

\def\etal{\emph{et al}\bmvaOneDot}
\def\vs{\emph{vs}\bmvaOneDot}

\DeclareMathOperator*{\argmax}{arg\,max}

\makeatletter
\renewcommand\paragraph{\@startsection{paragraph}{4}{\z@}%
  {.2em \@plus1ex \@minus.5ex}% .5em
  {-.5em}%
  {\normalfont\normalsize\bfseries\textcolor{bmv@captioncolor}}}
\makeatother

\begin{document}

\maketitle

\begin{abstract}
This paper strives to address image classifier bias, with a focus on both feature and label embedding spaces. Previous works have shown that spurious correlations from protected attributes, such as age, gender, or skin tone, can cause adverse decisions. To balance potential harms, there is a growing need to identify and mitigate image classifier bias.
First, we identify in the feature space a bias direction. We compute class prototypes of each protected attribute value for every class, and reveal an existing subspace that captures the maximum variance of the bias.
Second, we mitigate biases by mapping image inputs to label embedding spaces. Each value of the protected attribute has its projection head where classes are embedded through a latent vector representation rather than a common one-hot encoding.
Once trained, we further reduce in the feature space the bias effect by removing its direction.
Evaluation on biased image datasets, for multi-class, multi-label and binary classifications, shows the effectiveness of tackling both feature and label embedding spaces in improving the fairness of the classifier predictions, while preserving classification performance.
\end{abstract}

%-------------------------------------------------------------------------
\section{Introduction}
This paper strives to identify and mitigate biases present in image classifiers, with a focus on their feature and label embedding space. Adverse decisions from image classifiers can create discrimination against members of certain class of protected attribute, such as age, gender, or skin tone.
Buolamwini and Gebru~\cite{buolamwini2018gender} importantly show that face recognition systems misclassify subgroups with darker skin tones. This also applies to object recognition, where performance is higher for high-income communities~\cite{devries2019everyone} mainly located in Western countries~\cite{shankar2017geodiversity}.
Similarly problematic, current classifiers perpetuate and amplify current discrimination present in society~\cite{caliskan2017semantics,garg2018word}. For example, Kay~\etal~\cite{kay2015unequal} highlight the exaggeration of gender bias in occupations by image search systems.
These adverse decisions notably arise because image classifiers are prone to biases present in the dataset~\cite{geirhos2020shortcut}.
It is therefore essential to identify harmful biases in image representations and assess their effects on the classification predictions, as we do in this paper.

Addressing dataset biases is not enough, and classifier biases should also be addressed.
Zhao~\etal~\cite{zhao2017menshopping} importantly show that biases can actually be amplified during the image classifier training. Even when balancing a dataset for the protected attribute gender, image classifiers can still surprisingly amplify biases when making a prediction~\cite{wang2019balanced}.
This outcome emphasizes the importance of considering protected attributes during the training to avoid biased and adverse decisions.
A first approach is to perform \textit{fairness through blindness}, where the objective is to make the feature space blind to the protected attribute~\cite{zhang2018mitigating,alvi2018turning,hendricks2018women}.
An alternative is to perform \textit{fairness through awareness}, where the classifier label space is explicitly aware of the protected attribute label~\cite{dwork2012fairness}.
To better understand the effectiveness of these methods, Wang~\etal~\cite{wang2020towards} propose crucial benchmarks in biased image classification. They notably expose the shortcomings of these methods and show that a simple method with separate classifiers is more effective at mitigating biases.
Building on this line of work, this paper first identifies a bias direction in the feature space, and secondly address bias mitigation in both label and feature spaces.
Another important aspect concerns how to measure the fairness of image  classifiers. We borrow from the general fairness literature~\cite{beutel2017data,dwork2012fairness,hardt2016equality} to ensure that predictions are similar for all members of a protected attribute, which complements the benchmarks introduced by Wang~\etal~\cite{wang2020towards} on image classification bias.

\paragraph{Contributions.} 
Our main contribution is to demonstrate the importance of feature and label spaces for addressing image classifier bias.
First, we identify a bias direction in the feature space of common classifiers. We aggregate class prototypes to represent every class of each protected attribute value, and show a main direction to explain the maximum variance of the bias.
Second, we mitigate biases at both classification and feature levels. We introduce protected classification heads, where each head projects the features to a label embedding space specific to each protected attribute value. This differs from common classification, which usually considers a one-hot encoding for the label space~\cite{wang2020towards,saito2018maximum,luo2019taking}. For training, we derive a cosine softmax cross-entropy loss for multi-class, multi-label and binary classifications. Once trained, we apply in the feature space a bias removal operation to further reduce the bias effect.
Experiments
show the benefits on addressing classifier bias in both feature and label embedding spaces to improve fairness scores, while preserving the classification performance. The source code is available at: \href{https://github.com/twuilliam/bias-classifiers}{https://github.com/twuilliam/bias-classifiers}.

\section{Related Work}

\paragraph{Biases in word embeddings.}

Assessing the presence of biases in word embeddings, especially the gender bias, has received large attention given their wide range of applications within and beyond natural language processing.
The seminal and important work of Bolukbasi~\etal~\cite{bolukbasi2016man} reveals that the difference between female and male entities in word2vec~\cite{mikolov2013efficient} contains a gender bias direction. This shows that word2vec implicitly captures gender biases, which in return creates sexism in professional activities. Caliskan~\etal~\cite{caliskan2017semantics} further reveal that multiple human-like biases are actually present in word embeddings.
Even contextualized word embeddings~\cite{peters2018deep} are affected by a gender bias direction~\cite{zhao2019context}, which creates harmful risks~\cite{bender2021dangers}.
To mitigate such gender bias, Bolukbasi~\etal~\cite{bolukbasi2016man} propose a post-processing removal operation while Zhao~\etal~\cite{zhao2018genderneutral} derive regularizers to control the distance between relevant words during training.
It is important to note that biases cannot be removed entirely as they can still be recovered to some extent~\cite{gonen2019lipstick}. As such, methods mainly mitigate biases in models rather than producing debiased models.
Inspired by the literature on gender bias identification and mitigation in word embeddings, we pursue an analogous reasoning to show that biases are implicitly encoded in image classification models as well.

\paragraph{Biases in image datasets.}
As computer vision research relies heavily on datasets, they constitute a main source of biases.
Torralba and Efros~\cite{torralba2011unbiased} identify that datasets have a strong built-in bias as they only represent a narrow view of the visual world, leading models to rely on spurious correlations and produce detrimental predictions.
For fairness and transparency purposes, it becomes necessary to document the dataset creation~\cite{gebru2018datasheets,hutchinson2021towards}, as well as detecting the presence of potential biases and harms due to an unfair and unequal label sampling~\cite{shankar2017geodiversity,yang2020towards,birhane2021large,dixon2018measuring}. 
Towards this end, Bellamy~\etal~\cite{bellamy2018ai} and Wang~\etal~\cite{wang2020revisetool} propose metrics to measure biases, and actionable insights to mitigate them in a dataset.
Even though addressing biases when collecting a dataset is highly recommended, models can still produce unfair decisions~\cite{wang2019balanced}. In this paper, we focus on addressing image classifier bias.

\paragraph{Biases in image classifiers.}

% reveal
Searching for a representative subset of image examples provides visual explanations of biases~\cite{kim2016examples,stock2018convnets}. In this paper, we rather identify that such bias exists in the feature space in image classifiers.
% mitigate
To mitigate image classification bias, training with adversarial learning~\cite{goodfellow2014generative} makes the classifier blind to the protected attribute.
Reducing the gender bias can be achieved by forcing a model to avoid looking at people to produce a prediction~\cite{hendricks2018women,wang2019balanced}.
Blindness can also be achieved in the feature space by removing the variation of the protected attribute~\cite{alvi2018turning,zhang2018mitigating}.
Though, Wang~\etal~\cite{wang2020towards} illustrate that adversarial approaches tend to be detrimental as they decrease the performance by making image classifiers less discriminative. At the same time, non-adversarial approaches tend to amplify biases less, while performing well on image classification. Wang~\etal~\cite{wang2020towards} notably show that encoding the protected attribute into separate heads better mitigates biases.
We build on this literature and propose to mitigate biases at both classification and feature levels.

\paragraph{Biases benchmarking.}
No consensus exists (yet) in mitigating image classifier bias, which makes apple-to-apple comparisons complicated:
(a) benchmarks become no longer valid because datasets are taken down for ethical reasons~\cite{peng2021mitigating} (e.g., Racial faces in-the-wild~\cite{wang2019racial} derives from the problematic MS-Celeb-1M~\cite{guo2016MSCeleb1MAD}, and Diversity in Faces~\cite{merler2019diversity} has received complaints);
(b) datasets are introduced without benchmarks of debiasing methods (e.g., FairFace~\cite{karkkainen2021fairface} mainly evaluates commercial facial classification systems);
(c) related works come with differing evaluation settings (e.g., Wang~\etal~\cite{wang2019balanced} train MLP probes to measure model leakage).
While addressing algorithm bias in face verification~\cite{gong2020jointly,singh2020robustness,yin2019feature} is crucial, we focus on image classification~\cite{wang2019balanced,wang2020towards,kim2019learning,hwang2020exploiting}.
Therefore, we adopt in this paper the benchmarks introduced by Wang~\etal~\cite{wang2020towards} and Kim~\etal~\cite{kim2019learning} in multi-class, multi-label and binary classifications for their comprehensiveness and reproducibility.

\section{Identifying a Bias Direction}~\label{sec:reveal}
\vspace{-1.5em}

\paragraph{Problem formulation.}

We consider the task of image classification where every image $\bm{x}$ is assigned a label $y \in \mathcal{Y}$. For every image, there also exists a protected attribute value $v\in \mathcal{V}$, on which the classifier should not base its decision. In other words, classifiers should not discriminate against specific members of a protected attribute. In this paper, we consider discrete variables for protected attribute values, and limit the problem to binary values with $\mathcal{V}{=}\{0,1\}$. For example, we only consider the values \say{female} and \say{male} to describe the protected attribute \textit{gender}. It is important to note that this formulation is a simplification of the real world where protected attributes go beyond binary values, and are non-discrete.

Image classifiers are typically composed of a base encoder and a projection head. First, a base encoder $f(\cdot)$ extracts the feature representations of images $\bm{x}$. In our case, this corresponds to a convolutional network and results in $\bm{h}{=}f(\bm{x})$. Second, a projection head $g(\cdot)$ maps the features $\bm{h}$ to a discriminative space where a class is assigned. In our case, this corresponds to a linear projection, or a multilayer perceptron, and results in $\bm{z}{=}g(\bm{h})$ with $\bm{z} \in \mathbb{R}^M$. For example, in a one-hot encoding, $M$ equals the number of classes.

During training, we are given access to the protected attribute labels and can incorporate it in model formulations. We denote the triplet $(\bm{x}_i, y_i, v_i)$ as the $i$-th sample in the training set. During the evaluation, models only have access to the images. In this section, we show that common image classifiers -- that do not leverage protected attribute labels during training -- still implicitly encode their information in the feature space.

\begin{figure}[t]
\centering
\begin{minipage}{.48\textwidth}
\centering
\vspace{1.3em}
\hfill
\begin{overpic}[width=0.9\linewidth]{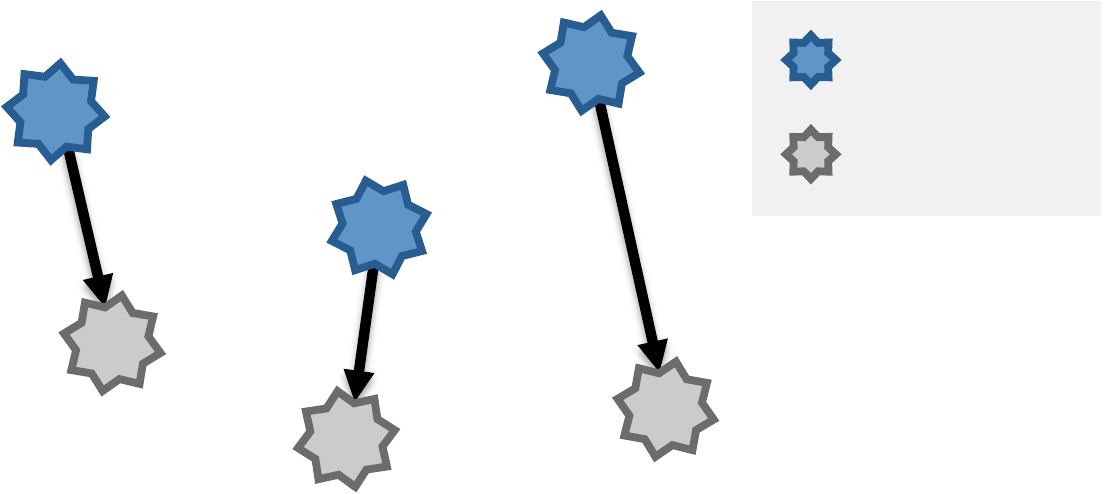}\label{fig:anyb}
\put(77.5,38){\footnotesize 0 \ding{221} color}
\put(77.5,28){\footnotesize 1 \ding{221} gray}

\put(0,42){$\bm{\mu}_{\textrm{cat}}^{0}$}
\put(5,3){$\bm{\mu}_{\textrm{cat}}^{1}$}

\put(27,32){$\bm{\mu}_{\textrm{dog}}^{0}$}
\put(27,-6){$\bm{\mu}_{\textrm{dog}}^{1}$}

\put(48,47){$\bm{\mu}_{\textrm{frog}}^{0}$}
\put(55,-3){$\bm{\mu}_{\textrm{frog}}^{1}$}

\put(-4,22){$\bm{\delta}_{\textrm{cat}}$}
\put(19,15){$\bm{\delta}_{\textrm{dog}}$}
\put(43,24){$\bm{\delta}_{\textrm{frog}}$}

\end{overpic}
\vspace{1.2em}
\caption{\textbf{2D toy visualization} of the feature space, where class prototypes $\bm{\mu}$ represent three categories with a color bias (\ding{88} \vs~{ \color{gray}\ding{88}}). A bias vector $\bm{\delta}$ is computed for every class.}
\label{fig:toy}
\end{minipage}\hfill
\begin{minipage}{.48\textwidth}
\centering
\vspace{-0.1em}
\subfigure[\centering $\bm{\Delta}$.]{
\includegraphics[width=0.5\linewidth]{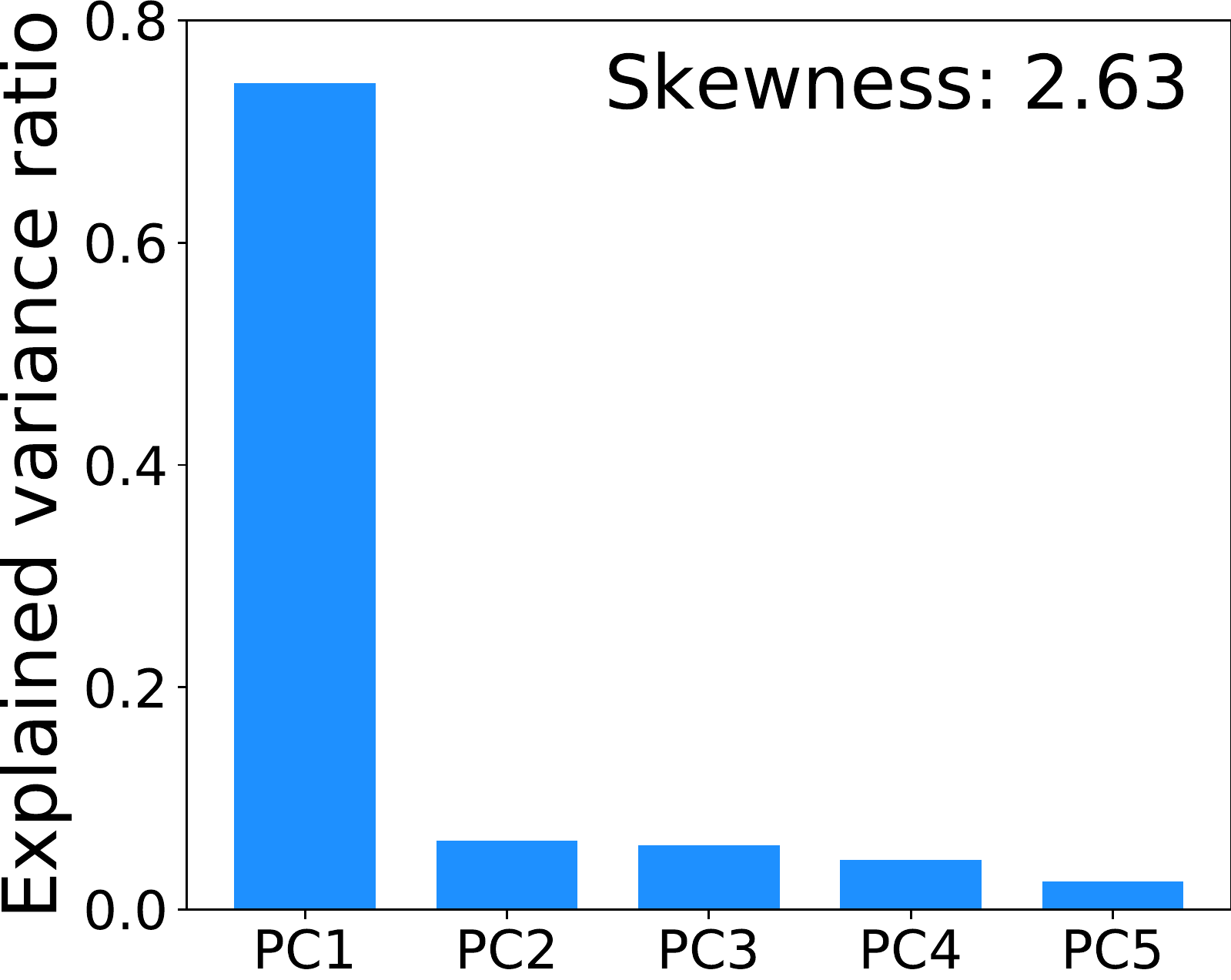}\label{fig:bias:a}}\hfill
\subfigure[\centering Random $\bm{\Delta}$.]{
\includegraphics[trim=28px 0 0 0, clip, width=0.47\linewidth]{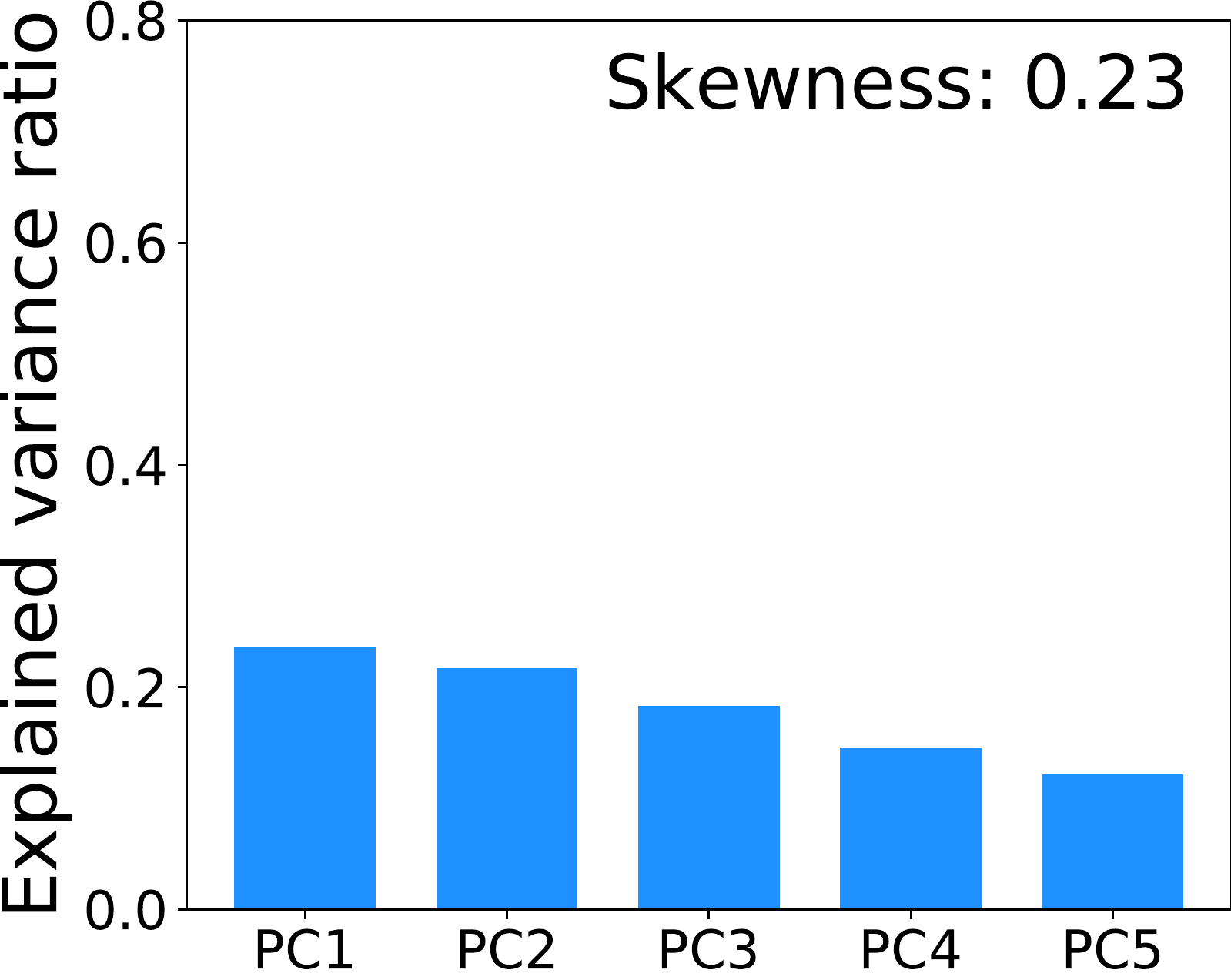}\label{fig:bias:b}}
\caption{\textbf{Bias direction} in the feature space. (a) The PCA of $\bm{\Delta}$ shows the maximum variance as the bias direction. (b) On a random $\bm{\Delta}$, the direction disappears and the explained variance is no longer skewed.
}
\label{fig:bias}
\end{minipage}
\end{figure}

\paragraph{Protected class prototypes.}
Once a model has been trained, we extract the features $\bm{h}$ from the training set. We then aggregate prototypes $\bm{\mu}_{y}^{v}$ for every class $y$ and specific to each protected attribute value $v$, coined as protected class prototypes. For example in Figure~\ref{fig:toy}, the class $y{=}\textrm{cat}$ has two prototypes in the feature space, one for $v{=}color$ images and one for $v{=}gray$. For any class $y$ with any protected attribute value $v$, we compute the protected class prototypes as their average representation in the feature space from the training set:
\begin{equation}
% \label{eq:py}
\bm{\mu}_{y}^{v} = \frac{1}{N_{y}^{v}} \sum_{i} \mathbb{I}[y_i=y \cap v_i=v] f(\bm{x}_i),
\end{equation}
where $N_{y}^{v}$ is the number of training images of class $y$ with protected attribute $v$, and $\mathbb{I}[\cdot]$ is the indicator function. Once all protected class prototypes are computed, we extract a subspace that captures the variance of the bias related to the protected attribute.

\paragraph{Bias direction.}
To identify a bias direction, we experiment with a standard convolutional network trained with a softmax cross-entropy loss on CIFAR-10S~\cite{wang2020towards}. This dataset provides a simple testbed to measure biases in images, as certain classes are skewed towards \textit{gray} images, while others are skewed towards \textit{color} images. Once trained, we aggregate the difference between class prototypes of each protected attribute value for every class:
\begin{equation}
\label{eq:delta}
\bm{\Delta} = \{\bm{\delta}_y | y \in \mathcal{Y}\} = \{\bm{\mu}_{y}^{1} - \bm{\mu}_{y}^{0} | y \in \mathcal{Y}\}.
\end{equation}
Note that for multi-label classification, we consider all binary labels to define $\mathcal{Y}$.
Figure~\ref{fig:bias:a} shows the principal component analysis (PCA) of $\bm{\Delta}$. When computing the ratio of explained variance of every principal component (PC), a main direction of variance appears. The first PC is more important than the others, which yields a high skewness. Figure~\ref{fig:bias:b} depicts the same analysis on a random $\bm{\Delta}$, where no main direction appears. Hence, there exists a subspace in the feature space where the bias information is maximized.

% \vspace{-0.25em}
\section{Mitigating Biases}
\vspace{-0.25em}

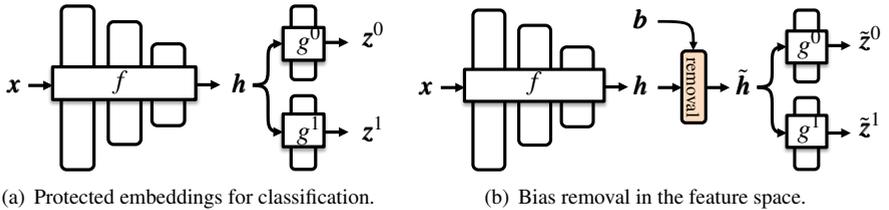
\begin{figure*}[t]\centering
\vspace{-0.5em}
\hfill
\subfigure[\centering Protected embeddings for classification.]{
\begin{overpic}[width=0.37\linewidth,trim=-14 0 -14 0,clip]{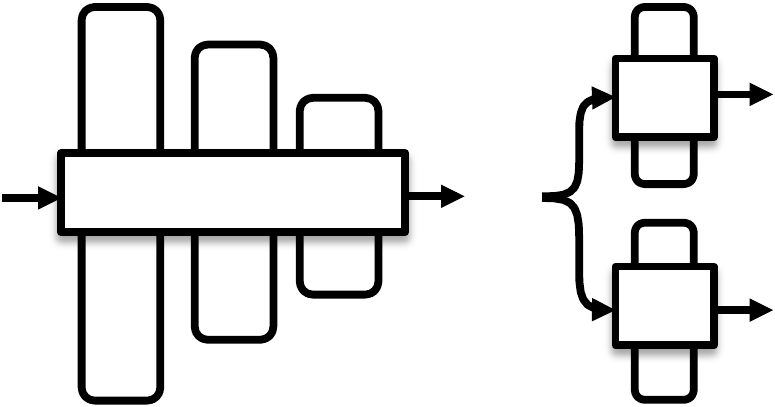}
\put(0,22){$\bm{x}$}
\put(62,22){$\bm{h}$}
\put(98,35){$\bm{z}^0$}
\put(98,9){$\bm{z}^1$}
\put(29,23){$f$}
\put(80,34){$g^0$}
\put(80,9){$g^1$}

\end{overpic}
}\hfill
\subfigure[\centering Bias removal in the feature space.]{
\begin{overpic}[width=0.42\linewidth]{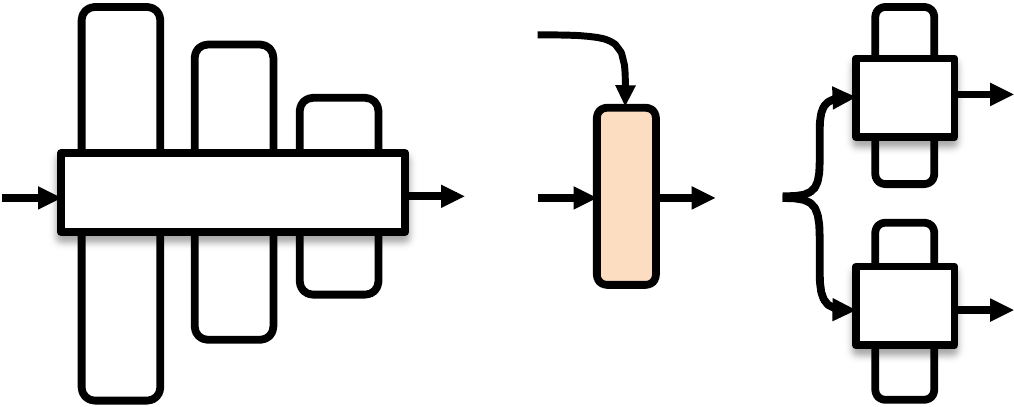}
\put(-5,19){$\bm{x}$}
\put(47,19){$\bm{h}$}
\put(72,19){$\bm{\tilde h}$}
\put(47,35){$\bm{b}$}
\put(60,28){\scriptsize\rotatebox{-90}{removal}}
\put(102,30){$\bm{\tilde z}^0$}
\put(102,9){$\bm{\tilde z}^1$}
\put(21,20){$f$}
\put(87,29){$g^0$}
\put(87,8.5){$g^1$}
\end{overpic}
}
\hfill\null
\caption{\textbf{Mitigating biases in classification predictions.} (a) For classification, we mitigate biases with protected label embeddings where each protected attribute value has its own space. (b) In the feature space, we include a removal operation of the bias direction $\bm{b}$ once the model has been trained, where $\bm{b}$ is computed from the training set.
}
\label{fig:model}
\vspace{-1em}
\end{figure*}

Figure~\ref{fig:model} illustrates our approach to mitigate biases in class predictions at both classification and feature levels. For the classification level, we create two protected label embedding spaces, one for each value of the binary protected attribute. For the feature level, we propose a bias removal operation once the model has been trained. The proposed method works for multi-class, multi-label and binary settings.

\paragraph{Protected label embeddings.}

We project features $\bm{h}$ into embedding spaces, one for each protected attribute value. This results in the embedding representation $\bm{z}^v=g^v(\bm{h}) \in \mathbb{R}^M$, where classification occurs.
During training, each projection head $g^v(\cdot)$ only sees samples from its assigned attribute value, which creates a protected embedding. By only seeing samples of one protected value, class boundaries are better separated~\cite{saito2018maximum}.

We further push these properties by relying on a cosine softmax cross-entropy loss for classification. $\bm{z}$ constitutes a discriminative embedding representation with semantic information about classes. This differs from related approaches in domain adaptation~\cite{saito2018maximum,luo2019taking} or bias mitigation~\cite{wang2020towards}, which also show the benefits of separate projection heads with a standard softmax but with a one-hot encoding label space. Below we derive a cosine softmax with protected embeddings for both multi-class, multi-label and binary classifications.

\paragraph{Multi-class classification} assigns a label $y \in \mathcal{Y}$ to an image $\bm{x}$. We introduce a protected weight matrix $\bm{W}^v \in \mathbb{R}^{|\mathcal{Y}| \times M}$, where $M$ is the size of the embedding space and $v \in \mathcal{V}$ is the protected attribute value. Every row $\bm{W}_{y,:}^v$ acts as a latent real-valued semantic representation for every class $y$ of each protected attribute $v$. The objective is then to maximize the cosine similarity, denoted as \say{sim}, between an embedding representation $\bm{z}^v$ and its corresponding weight representation. This results in the probabilistic model:
\begin{equation}
% \label{eq:py}
p(y|\bm{z}^v,v) = \frac{\exp\Big(\textrm{sim}(\bm{W}_{y,:}^v, \bm{z}^v)/\tau\Big)}{\sum_{y' \in \mathcal{Y}}\exp\Big(\textrm{sim}(\bm{W}_{y',:}^v, \bm{z}^v)/\tau \Big)},
\end{equation}
where $\tau$ is a temperature scaling hyper-parameter. For training, we minimize the cross-entropy loss over the training set of size $N$:
$\mathcal{L} = - \frac{1}{N} \sum_{i}^N \sum_{v' \in \mathcal{V}} \mathbb{I}[v_i=v'] \log p(y_i|\bm{x}_i, v_i)$. During inference, the attribute value label is not present. Thus, we perform an ensemble prediction over both heads to predict $\hat{y} = \argmax_y \sum_{v' \in \mathcal{V}} p(y|\bm{x},v')$.

\paragraph{Multi-label classification} assigns multiple binary labels $\bm{y}$ to an image $\bm{x}$. This typically occurs when we want to predict the presence of multiple binary attributes in an image. We denote as $y^{(c)} \in \{0, 1\}$ the label of attribute $c$. Similar to multi-class classification, we introduce a protected weight matrix $\bm{W}^{v,c} \in \mathbb{R}^{2 \times M}$ where the two rows correspond to the absence and presence of attribute $c$ for protected attribute $v$. The resulting probabilistic model is:
\begin{equation}
\label{eq:multilabel}
p(y^{(c)}|\bm{z}^v,v) = \frac{\exp\Big(\textrm{sim}\left(\bm{W}_{y,:}^{v,c}, \bm{z}^v)\right)/\tau\Big)}{\sum_{y' \in \{0, 1\}}\exp\Big(\textrm{sim}(\bm{W}_{y',:}^{v,c}, \bm{z}^v)/\tau \Big)},
\end{equation}
which corresponds to a classifier for two classes. Compared with a binary classifier with a sigmoid function, the softmax function offers more flexibility for the model to represent the negatives. We minimize the cross-entropy loss over all $C$ attributes of the training set of size $N$:
$\mathcal{L} = - \frac{1}{N\cdot C}\sum_{i=1}^{N}\sum_{c=1}^{C}\sum_{v' \in \mathcal{V}} \mathbb{I}[v_i=v'] \log p(y_i^c|\bm{x}_i, v_i)$.
During inference, we also perform an ensemble prediction to compute the probability score for the presence of every attribute $\hat{y}^c = \sum_{v' \in \mathcal{V}} p(y^{(c)}=1|\bm{x},v')$.
Binary classification is a special case where $C{=}1$.

\paragraph{Bias removal in the feature space.}
Once trained, we perform the same analysis as in Section~\ref{sec:reveal} where we collect protected class prototypes in the feature space from the training set and also apply a principal component analysis on their differences $\bm{\Delta}$. We refer to the direction of the first principal component of $\bm{\Delta}$ as $\bm{b}$. Following Bolukbasi~\etal~\cite{bolukbasi2016man}, we first project features $\bm{h}$ on the bias direction $\bm{b}$ to obtain $\bm{h}_{\bm{b}}$. Then, we neutralize the bias effect by removing  $\bm{h}_{\bm{b}}$ from the features $\bm{h}$, resulting in the mitigated features $\bm{\tilde h}$. Mathematically, this bias removal operation corresponds to: $\bm{\tilde h} = \bm{h} - \bm{h}_{\bm{b}} = \bm{h} - \frac{\bm{h}\cdot \bm{b}}{\lVert \bm{b} \lVert} \frac{\bm{b}}{\lVert \bm{b} \lVert}.$
Once $\bm{\tilde h}$ is computed, we can further feed it to each head to get the mitigated protected embeddings $\bm{\tilde z}^{v}{=}g^v(\bm{\tilde h})$.

\paragraph{Relation with \textit{Domain Independent}~\cite{wang2020towards}.}
Our proposed method builds on the observation from Wang~\etal~\cite{wang2020towards} that separate classification heads improve the fairness of the predictions.
We differ by demonstrating how feature and label spaces also matter for addressing biases. We find the feature space implicitly encodes a bias direction (Section~\ref{sec:reveal}) and we derive a bias removal operation to reduce its influence. As distances matter in the feature space, this motivates us to switch from a one-hot encoding to a real-valued vector representation for the label space, where classification now occurs through a cosine embedding softmax. 

\vspace{-0.5em}
\section{Experiments}
\vspace{-0.25em}
\subsection{Fairness Metrics}

\paragraph{Bias amplification} measures whether spurious correlations present in the dataset have been amplified by the model during training~\cite{zhao2017menshopping}.
Following Zhao~\etal~\cite{zhao2017menshopping}, the bias amplification score corresponds to:
$\frac{1}{|\mathcal{Y}|} \sum_{v \in \mathcal{V}} \sum_{y\in \mathcal{Y}} \mathbb{I}_{s(y, v) > \frac{1}{|\mathcal{V}|}}
\frac{P^{v}_y}{P^0_y + P^1_y} - s(y, v)$,
where $P^{v}_y$ is the number of images positive for class $y$ with a protected attribute $v$ predicted by the model, and $s(y, v){=}N^{v}_y/ (N^0_y + N^1_y)$ is the ratio of training images $N^{v}_y$ of class $y$ with a protected attribute $v$. Intuitively, the score should be as low as possible: a positive value indicates a bias amplification while a negative value indicates a bias reduction.
When training and testing sets are not \textit{i.i.d.}, we follow Wang~\etal~\cite{wang2020towards} and compute:
$\frac{1}{|\mathcal{Y}|}  \sum_{y \in \mathcal{Y}} \frac{\max (P_y^0, P_y^1)}{P_y^0 + P_y^1} - 0.5$.

\paragraph{Demographic parity} assesses the independence between a prediction $\hat y$ and a protected attribute $v$ such that $p(\hat y {=} y'|v{=}0) {=} p(\hat y {=} y'|v{=}1)$~\cite{hardt2016equality,dwork2012fairness}.
Following Beutel~\etal~\cite{beutel2017data}, a statistical parity difference score is derived:
$\frac{1}{|\mathcal{Y}|}\sum_{y\in \mathcal{Y}} \left| \frac{\textrm{TP}^1_y + \textrm{FP}^1_y}{N^1} - \frac{\textrm{TP}^0_y + \textrm{FP}^0_y}{N^0}\right|$,
where $\textrm{TP}^v_y$ and $\textrm{FP}^v_y$ are the number of true positives and false positives of class $y$ with protected attribute $v$, and $N^v$ is the number of images with protected attribute $v$ in the evaluation set. When the score tends to zero, the model makes the same rate of predictions for class $y'$ regardless of the protected attribute value.

\paragraph{Equality of opportunity} assesses the conditional independence on a particular class $y'$ between a prediction $\hat y$ and a protected attribute $v$ such that $p(\hat y = y'| y = y', v=0) = p(\hat y = y'| y = y', v=1)$~\cite{hardt2016equality}.
Following Beutel~\etal~\cite{beutel2017data}, a difference of equality of opportunity score is derived:
$\frac{1}{|\mathcal{Y}|}\sum_{y\in \mathcal{Y}}\left| \frac{\textrm{TP}^1_y}{\textrm{TP}^1_y+\textrm{FN}^1_y} - \frac{\textrm{TP}^0_y}{\textrm{TP}^0_y+\textrm{FN}^0_y}\right|$,
where $\textrm{FN}^v_y$ is the number of false negatives of class $y$ with protected attribute $v$. When the score tends to zero, the model classifies images as class $y'$ correctly regardless of the protected attribute value.

\paragraph{Equalized odds}
assesses the conditional independence on any class $y'$ between a prediction $\hat y$ and a protected attribute $v$ such that $p(\hat y = y'| y = y, v=0) = p(\hat y = y'| y = y, v=1)$~\cite{hardt2016equality}.
Following Bellamy~\cite{bellamy2018ai}, a difference of equalized odds score is derived:
$0.5\cdot (|FPR_y^1 - FPR_y^0| + |TPR_y^1 - TPR_y^0|)$,
where $FPR_y^v$ is the false positive rate of class $y$ with protected attribute $v$ and $TPR_y^v$ is the true positive rate. When the score tends to zero, the model exhibits similar true positive and false positive rates for both protected attribute values.

\subsection{Multi-class Classification}

\paragraph{Setup.}
We evaluate multi-class classification on the CIFAR-10S dataset~\cite{wang2020towards}, which is a biased version of the original CIFAR-10 dataset~\cite{krizhevsky2009learning}. A color bias is introduced in the training set, where 5 classes contain 95\% gray images and 5\% color images, and conversely for the 5 other classes.
Visual examples for every class in their dominant color bias are present in the appendix.
This creates simple spurious correlations that still affect common classifiers. Two versions of the testing set are considered: one with only gray images and another one with only color images. Although this breaks the \textit{i.i.d.} assumption between training and testing sets, it allows the assessment of the color bias in a controlled manner.
We report the per-class accuracy over 5 runs.
We rely on ResNet18~\cite{he2016deep} as the encoding function $f$ and set each projection function $g^v$ as a fully-connected layer of size $M{=}128$ followed by a linear activation. Training is done from scratch with stochastic gradient descent with momentum~\cite{sutskever2013importance} for 200 epochs, and the following hyper-parameters: learning rate of 0.1 with a momentum of 0.9, batch size of 128, weight decay of 5e-4, and temperature of 0.1. The learning rate is reduced by a factor 10 every 50 epochs.
Note that this setup is identical for all models we compare with, as benchmarked by Wang~\etal~\cite{wang2020towards}.

\paragraph{Results.}

\begin{table*}[t]
\centering
\tablestyle{2pt}{.95}
\resizebox{0.98\linewidth}{!}{
\begin{tabular}{lllrrrrr}
\toprule
Model & Loss & Acc. {\scriptsize(\%,$\uparrow$)} & Bias {\scriptsize($\downarrow$)} &  Parity {\scriptsize(\%,$\downarrow$)} & Opp. {\scriptsize(\%,$\downarrow$)} & Odds {\scriptsize(\%,$\downarrow$)} \\
% \midrule
\cmidrule(lr){1-1}\cmidrule(lr){2-2}\cmidrule(lr){3-3} \cmidrule(lr){4-7}
{\sc Baseline} & N-way softmax & 88.5{\scriptsize$\pm$0.3} & 0.074{\scriptsize$\pm$0.003} & 2.90{\scriptsize$\pm$0.11} & 13.07{\scriptsize$\pm$0.37} & 7.19{\scriptsize$\pm$0.21} \\
{\sc Oversampling} & N-way softmax & 89.1{\scriptsize$\pm$0.4} & 0.066{\scriptsize$\pm$0.002} & 2.77{\scriptsize$\pm$0.67} & 12.58{\scriptsize$\pm$0.19} & 6.91{\scriptsize$\pm$0.11}\\
\midrule
\multirow{2}{*}{{\sc Adversarial}} & w/ confusion~\cite{alvi2018turning,tzeng2015simultaneous}  & 83.8{\scriptsize$\pm$1.1} & 0.101{\scriptsize$\pm$0.007} & 4.14{\scriptsize$\pm$0.28} & 16.71{\scriptsize$\pm$1.37} &
9.28{\scriptsize$\pm$0.73} \\
& w/ \(\nabla\) rev. proj. \cite{ganin2016domain} & 84.1{\scriptsize$\pm$1.0} & 0.094{\scriptsize$\pm$0.011} & 3.60{\scriptsize$\pm$0.46} & 14.13{\scriptsize$\pm$1.43} & 
7.89{\scriptsize$\pm$0.81} \\
\midrule 
{\sc Domain Discriminative} & joint ND-way softmax & 90.3{\scriptsize$\pm$0.5} & 0.040{\scriptsize$\pm$0.002} & 1.65{\scriptsize$\pm$0.06} & 7.27{\scriptsize$\pm$0.32} &
4.02{\scriptsize$\pm$0.17} \\
\midrule
{\sc Domain Independent} & N-way softmax$\times$D & \textbf{92.0}{\scriptsize$\pm$0.1} & \textbf{0.004}{\scriptsize$\pm$0.001} & 0.20{\scriptsize$\pm$0.04} & 1.07{\scriptsize$\pm$0.22} &
0.59{\scriptsize$\pm$0.12} \\
\midrule
\multirow{1}{*}{\textit{This paper}} & \multirow{1}{*}{N-way cos softmax$\times$D} & 91.5{\scriptsize$\pm$0.2} & \textbf{0.004}{\scriptsize$\pm$0.000} &\textbf{0.15}{\scriptsize$\pm$0.01} & \textbf{0.83}{\scriptsize$\pm$0.12}  &
\textbf{0.46}{\scriptsize$\pm$0.07} \\
\bottomrule
\end{tabular}
}
\vspace{0.1in}
\caption{\textbf{Multi-class classification comparison} on $N{=}10$ classes of CIFAR-10S.
Despite a small loss in the accuracy score, our proposed approach with a cosine softmax, rather than a common softmax as in {\sc Domain independent}, improves the fairness of the model in multi-class classification.}
\label{table:cifar:sota}
\vspace{-1em}
\end{table*}

Table~\ref{table:cifar:sota} compares our method with four other approaches.
{\sc Baseline} is a standard model trained with an N-way softmax while {\sc Oversampling} balances out the training by sampling more often underrepresented values of the protected attribute.
{\sc Adversarial} blinds the feature space to the protected attribute. This is achieved either with a uniform confusion loss~\cite{alvi2018turning,tzeng2015simultaneous} or a gradient reversal layer~\cite{ganin2016domain}.
{\sc Domain discriminative} makes the classification aware of the protected attribute label by assigning a class for every category and protected attribute pair~\cite{dwork2012fairness}.
{\sc Domain independent} creates two classification heads, one head for each value of the protected attribute~\cite{wang2020towards}. 
Reported accuracy and bias amplification scores correspond to Wang~\etal~\cite{wang2020towards}, while we reproduce their experiments from the source code for the demographic parity, equality of opportunity, and equalized odds scores.

Our proposed approach improves upon the other alternatives in the fairness scores. Only in the accuracy metric our model yields slightly lower results compared with {\sc Domain independent}. This shows that there might exist a trade-off between the downstream task and the fairness of the classifier, as improving both remains challenging.
It is interesting that {\sc Adversarial} produces worse results than simple methods such as {\sc Baseline} or {\sc Oversampling}. As {\sc Adversarial} blurs the distinction between both protected attribute values, it also alters the class boundaries, which makes the model less discriminative.
{\sc Domain discriminative} achieves a lower performance than our model and {\sc Domain independent}. This highlights the importance of separating the classification heads for each protected attribute value.
Overall, our proposed approach with a cosine softmax, rather than a common softmax as in {\sc Domain independent}, reduces the bias direction in the feature space (see appendix) and improves the fairness in multi-class classification.

\subsection{Multi-label Classification}
% \vspace{-.3em}
\paragraph{Setup.}

We evaluate multi-label classification on the \say{Align and Cropped} split of the CelebA dataset~\cite{liu2015faceattributes}, which contains 202,599 face images labeled with 40 binary attributes. Following Wang~\etal~\cite{wang2020towards}, we consider the gender as the protected attribute and train models to predict the other 39 attributes. 
Visual examples of attributes with a high gender skewness are presented in the appendix.
During the testing phase, only 34 attributes are considered as the other 5 do not contain both genders. We report the weighted mean average (mAP) precision across the selected attributes. Every positive man image is weighted by $(N_m + N_w)/(2N_m)$ while every positive woman image by $(N_m + N_w)/(2N_w)$, where $N_m$ and $N_w$ are the man and woman image counts in the test set. This weighting ensures a balanced representation of both genders in the evaluation of every attribute.

We rely on ResNet50~\cite{he2016deep} pre-trained on ImageNet~\cite{ILSVRC15} as the encoding function $f$. We remove the final classification layer and replace it with two fully-connected layers (one for each protected attribute $v$) of size $M{=}128$ followed by a linear activation as the projection function $g^v$. Training is done with stochastic gradient descent with momentum~\cite{sutskever2013importance}, and the following hyper-parameters: learning rate of 0.1 with a momentum of 0.9, batch size of 32, and temperature of 0.05. The best model is selected according to the weighted mAP score on the validation set.
Compared with the benchmarks introduced by Wang~\etal~\cite{wang2020towards}, our model training only differs by the optimizer, as we notice some overfitting issues when using Adam~\cite{kingma2014adam}. The backbone and the rest of the hyper-parameters are similar.

\begin{table}[t]
\begin{minipage}{.49\linewidth}
\centering
    \tablestyle{2pt}{.95}
\resizebox{0.98\linewidth}{!}{
\begin{tabular}{lcrrrr}
\toprule
Loss & mAP & Bias & Parity & Opp. & Odds \\
\cmidrule(lr){1-1}\cmidrule(lr){2-2}\cmidrule(lr){3-6}
N sigmoids $\times$ D & 75.4 & -0.039 & 17.74 & 14.87 & 9.19 \\
N cos sigmoids  $\times$ D & 75.5 & 0.001 & 11.63 & 10.29 & 5.79 \\
+ bias removal & 74.7 & -0.020 & 7.43 & 7.00 & \textbf{4.00} \\

\midrule
N cos softmax $\times$ D & \textbf{76.3} & -0.006 & 11.97 & 10.18 & 6.06 \\
+ bias removal & 75.3 & \textbf{-0.041} & \textbf{6.71} & \textbf{6.73}  & 4.10 \\
\bottomrule
    \end{tabular}
    }
    \vspace{0.1in}
    \caption{\textbf{Label space} comparison on CelebA.
    An embedding learned with a cosine similarity improves the fairness upon common sigmoids. A softmax with bias removal in the feature space further improves fairness.
    }
    \label{table:celeba:loss}
\end{minipage}
\hfill
\begin{minipage}{.49\linewidth}
\centering
    \centering
    \tablestyle{2pt}{.95}
    \vspace{-0.11in}
    \resizebox{0.98\linewidth}{!}{
    \begin{tabular}{llrrrrr}
\toprule
\multirow{2}{*}{Embedding} & \multirow{2}{*}{\shortstack[l]{Cos\\softmax}} & \multirow{2}{*}{mAP} & \multirow{2}{*}{Bias} & \multirow{2}{*}{Parity} & \multirow{2}{*}{Opp.} & \multirow{2}{*}{Odds} \\
\\
\cmidrule(lr){1-1}\cmidrule(lr){2-2}\cmidrule(lr){3-3}\cmidrule(lr){4-7}
Single & N & 74.5 & -0.039 & 10.65 & 14.02 & 7.77\\
Single & N $\times$ D & 67.7 & \textbf{-0.070} & 19.26 & 21.02 & 13.54\\
\midrule
Protected & N $\times$ D & \textbf{75.3} & -0.041 & \textbf{6.71} & \textbf{6.73} & \textbf{4.10}\\
\bottomrule\\
    \end{tabular}}
    \vspace{0.1in}
    \caption{\textbf{Single \vs protected embedding} comparison on CelebA. Separating the gender information into protected heads results in an increased classification and fairness performance over a single head.}
    \vspace{-0.1in}
    \label{table:celeba:emb}
\end{minipage}
% \vspace{-0.5em}
\end{table}

\begin{table*}[t]
    \centering
    \tablestyle{2pt}{.95}
    \resizebox{0.98\linewidth}{!}{
    \begin{tabular}{llrrrrr}
\toprule
Model & Loss & mAP {\scriptsize(\%,$\uparrow$)} & Bias {\scriptsize($\downarrow$)} &  Parity {\scriptsize(\%,$\downarrow$)} & Opp. {\scriptsize(\%,$\downarrow$)} & Odds {\scriptsize(\%,$\downarrow$)} \\
% \hline
\cmidrule(lr){1-1}\cmidrule(lr){2-2}\cmidrule(lr){3-3} \cmidrule(lr){4-7}
{\sc Baseline}  & N sigmoids & 74.7 & 0.010 & 23.32 & 24.34 & 14.28\\
\midrule
{\sc Adversarial}
&  w/ confusion~\cite{alvi2018turning,tzeng2015simultaneous}
& 71.9 & 0.019 & 23.73 & 28.66 & 16.69\\
\midrule
{\sc Domain Discriminative} & ND sigmoids & 73.8 & 0.007 & 22.34 & 25.35 & 14.69\\ 
\midrule
{\sc Domain Independent} & N sigmoids $\times$ D & 
\textbf{75.4} & -0.039 & 17.74 & 14.87 & 9.19\\
\midrule
\multirow{1}{*}{\textit{This paper}} & N cos softmax  $\times$ D &  75.3 & \textbf{-0.041} & \textbf{6.71} & \textbf{6.73} & \textbf{4.10} \\
\bottomrule
    \end{tabular}}
    \vspace{0.1in}
    \caption{\textbf{Multi-label classification comparison} of $N{=}34$ attributes in CelebA.
    Despite a small loss in the mAP score, our proposed embedding -- learned with a cosine softmax rather than a common softmax with one-hot encoding as in {\sc Domain independent} -- improves the fairness of the model in multi-label classification.}
    \label{table:celeba:sota}
    \vspace{-0.5em}
\end{table*}

\paragraph{Label space.}

Table~\ref{table:celeba:loss} compares the different formulations of the label embedding space.
Relying on a real-valued embedding space learned with a cosine similarity function improves the fairness of the predictions compared with the common one-hot representation. Labels now correspond to a real-valued vector instead of a binary value, which enables a distributed class representation.
Switching to a softmax function instead of a sigmoid provides a weight representation for negatives, which in return helps the classification performance.
The benefit of negative representations is further highlighted when applying the bias removal operation in the feature space, even though a small drop in the classification score occurs.
Overall, learning an embedding with a softmax cross-entropy, plus the bias removal, preserves the performance of the downstream task while improving the fairness of the predictions.

\paragraph{Single \vs protected embeddings.}

Table~\ref{table:celeba:emb} assesses the importance of having protected embeddings, with one projection function $g^v$ for each value $v$ of the protected attribute gender. We evaluate the single head setting with and without the protected attribute label in the loss function. When the protected attribute information is available, we basically have two cosine softmax losses, one for each value. Mixing the two losses in one single head is detrimental to the performance as the model gets confused on where to project the inputs in the embedding space. Protected embeddings better separate the gender information for the classification of every attribute as illustrated by the improved performance, and fairness scores overall.

\paragraph{Results.}

Table~\ref{table:celeba:sota} compares our model with four other approaches, similarly to the comparison in Table~\ref{table:cifar:sota}.
Reported mAP and bias amplification scores correspond to Wang~\etal~\cite{wang2020towards}, while we reproduce their experiments to measure demographic parity, equality of opportunity, and equalized odds scores.
Our proposed approach yields the fairer scores across all evaluated models. And similar to multi-class classification, we also notice a small drop in the downstream task when measuring the mAP.
The {\sc Adversarial} produces again the worst results across all metrics. This indicates that current methods applying an adversarial training remove more information than the bias, which is detrimental for both the downstream task and the fairness of the model.
{\sc Domain discriminative} and {\sc Baseline} result in a similar performance.
Interestingly, a trade-off between the mAP and fairness scores is also present in {\sc Domain independent}.
Our proposed approach improves over {\sc Domain independent} in the fairness scores by a large margin.
Mitigating the bias in both feature and label embedding spaces is then preferred over methods that only address one of the two.

\begin{table}[t]
\centering
\vspace{-0.25em}
\hfill
\subfigure[Gender prediction (age protected)]{
\tablestyle{6pt}{.95}
\begin{tabular}{lcccc}
\toprule
\multirow{2}{*}{Method} & \multicolumn{2}{c}{Trained on \textit{EB1}} & \multicolumn{2}{c}{Trained on \textit{EB2}}\\
& \textit{EB2} & \textit{Test} & \textit{EB1} & \textit{Test} \\
\cmidrule(lr){1-1}\cmidrule(lr){2-3}\cmidrule(lr){4-5}
{\sc Baseline} & 59.86 & 84.42 & 57.84 & 69.75 \\ 
Alvi~\etal~\cite{alvi2018turning} & 63.74 & 85.56 & 57.33 & 69.90 \\ 
Kim~\etal~\cite{kim2019learning} &  68.00 & 86.66 & 64.18 & 74.50 \\
\hline
\textit{This paper} & \textbf{70.85} & \textbf{88.73} & \textbf{80.59} & \textbf{83.65}  \\
\bottomrule
\end{tabular}
}
\quad
\subfigure[Age prediction (gender protected)]{
\tablestyle{2pt}{.95}
\begin{tabular}{cccc}
\toprule
\multicolumn{2}{c}{Trained on \textit{EB1}} & \multicolumn{2}{c}{Trained on \textit{EB2}}\\
\textit{EB2} & \textit{Test} & \textit{EB1} & \textit{Test} \\
\cmidrule(lr){1-2}\cmidrule(lr){3-4}
54.30 & 77.17 & 48.91 & 61.97 \\
\textbf{66.80} & 75.13 & 64.16 & 62.40 \\
54.27 & 77.43 & 62.18 & 63.04 \\
\hline
35.93 & \textbf{77.67} & \textbf{65.90} & \textbf{73.08} \\
\bottomrule
\end{tabular}
}\hfill\null
\caption{\textbf{Binary classification comparison} on IMDB face dataset.
Our formulation of the label embedding space improves the binary classification accuracy (\%) with an extreme bias over methods that impose an invariance to the protected attribute in the feature space.
}
\label{tab:bin}
\vspace{-0.5em}
\end{table}

\paragraph{Binary classification.}
We evaluate binary classification on the \say{cropped} split of the IMDB face dataset~\cite{Rothe2018IJCV}.
Following Kim~\etal~\cite{kim2019learning}, we create three sets with an extreme bias:
\textit{EB1} comprises women $\leq$ 29 years old (yo) and men $\geq$ 40 yo;
\textit{EB2} has women $\geq$ 40 yo and men $\leq$ 29 yo;
and \textit{Test} has women and men $\leq$ 29 yo and $\geq$ 40 yo.
They contain 36,004, 16,800 and 13,129 face images of celebrities.
Similar to Kim~\etal~\cite{kim2019learning}, we learn to predict the gender with age as a protected attribute (and conversely), and
rely on ResNet18~\cite{he2016deep} pre-trained on ImageNet~\cite{ILSVRC15} as the encoding function $f$. We add a fully-connected layer of size $M{=}128$ with linear activation for each projection function $g^v$.
Training is done with stochastic gradient descent with momentum~\cite{sutskever2013importance}, and a learning rate of 0.1 with momentum of 0.9 and an exponential decay of 0.999, batch size of 128, and temperature of 0.1.
Given the extreme bias, we update both protected heads instead of only one as done previously.

Table~\ref{tab:bin} compares our model with three other approaches.
{\sc Baseline} is also a standard model trained with binary cross-entropy. Both Alvi~\etal~\cite{alvi2018turning} and Kim~\etal~\cite{kim2019learning} mitigate the extreme bias by making the feature space invariant to the protected attribute. Kim~\etal~\cite{kim2019learning} rely on an adversarial formulation~\cite{ganin2016domain,chen2016infogan}, improving over Alvi~\etal~\cite{alvi2018turning}.
Given the binary classification setting, we did not apply a bias removal operation, as a PCA on two samples is not pertinent. Still, our formulation of the label space improves the performance in both the gender and age settings. Only when predicting age and training on \textit{EB1}, our model struggles a bit as it tends to overfit quickly.
This binary classification comparison further confirms that simpler alternatives to adversarial losses can better mitigate biases in image classifiers.

\vspace{-1em}
\section{Conclusion}
\vspace{-0.5em}

Reducing the effect of adverse decisions involves the identification and mitigation of biases within model representations.
In this paper, we focus on biases coming from binary protected attributes.
First, we identify a direction in the feature space of common image classifiers, where the first principal component of the difference of protected class prototypes captures bias variation.
Second, building on this observation, we mitigate bias with protected projection heads that learn a label embedding space for each protected attribute value. This formulation trained with a cosine softmax cross-entropy loss improves the fairness in multi-class, multi-label and binary classifications compared with a common one-hot encoding.
Removing the bias direction in the feature space reduces even further the bias effect on the classifier predictions.
Overall, addressing image classifier bias on both feature and label spaces improves the fairness of predictions, while preserving the classification performance.

\bibliography{egbib}

\clearpage

\appendix

~

\begin{raggedright}%^
{\LARGE\bfseries\sffamily\color{bmv@sectioncolor} Feature and Label Embedding Spaces Matter in Addressing Image Classifier Bias\par}%^
\vskip 1.5em%^
\end{raggedright}%^

\begin{center}
\sffamily\normalsize Supplementary material

\end{center}

\par\smallskip\hrule

\smallskip

\begin{figure}[h]\centering
\vspace{-0.5em}
\hfill
\subfigure[\centering Before.]{
\includegraphics[width=0.32\linewidth]{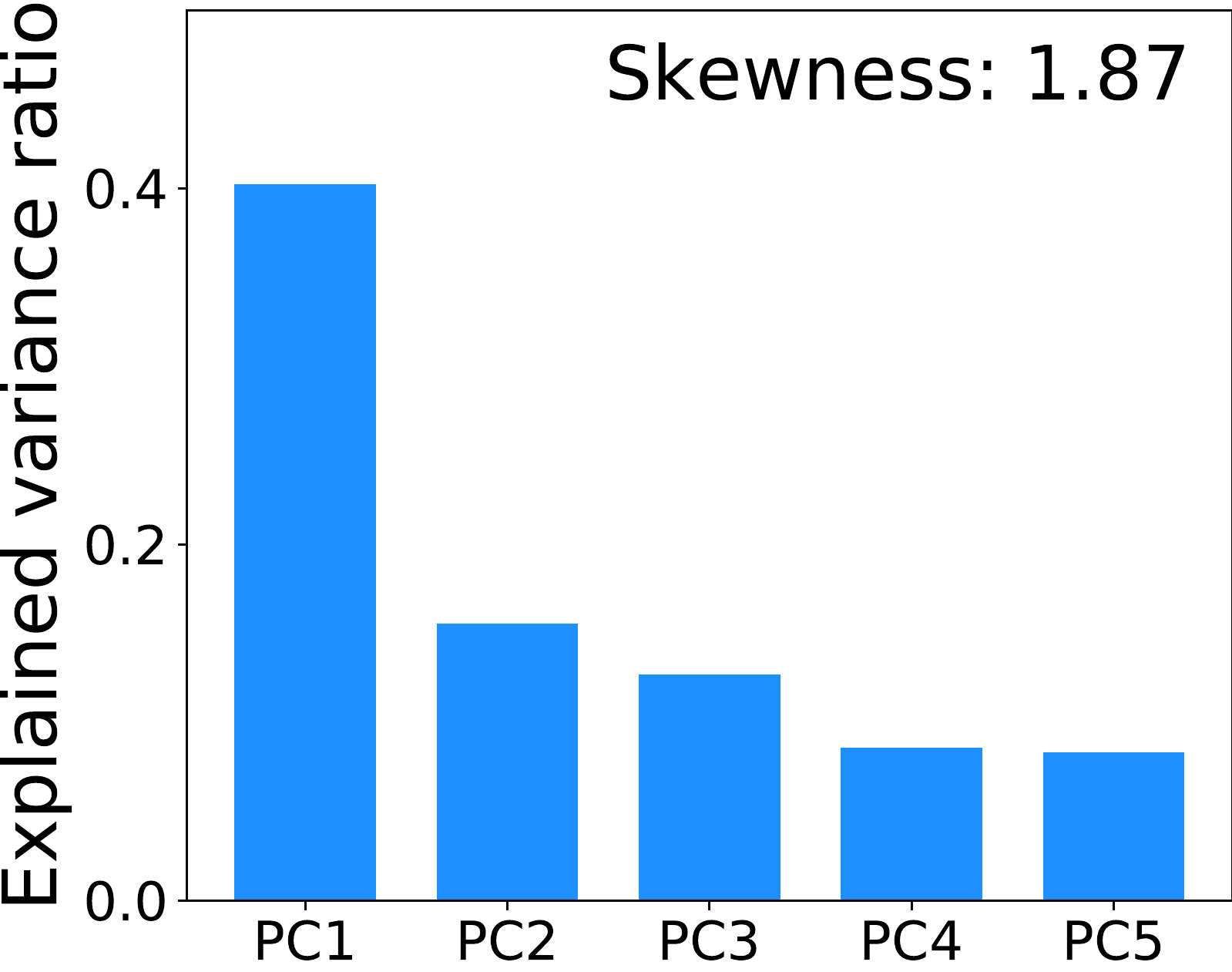}\label{fig:biasremoval:a}}\hfill
\subfigure[\centering After.]{
\includegraphics[trim=28px 0 0 0, clip, width=0.3\linewidth]{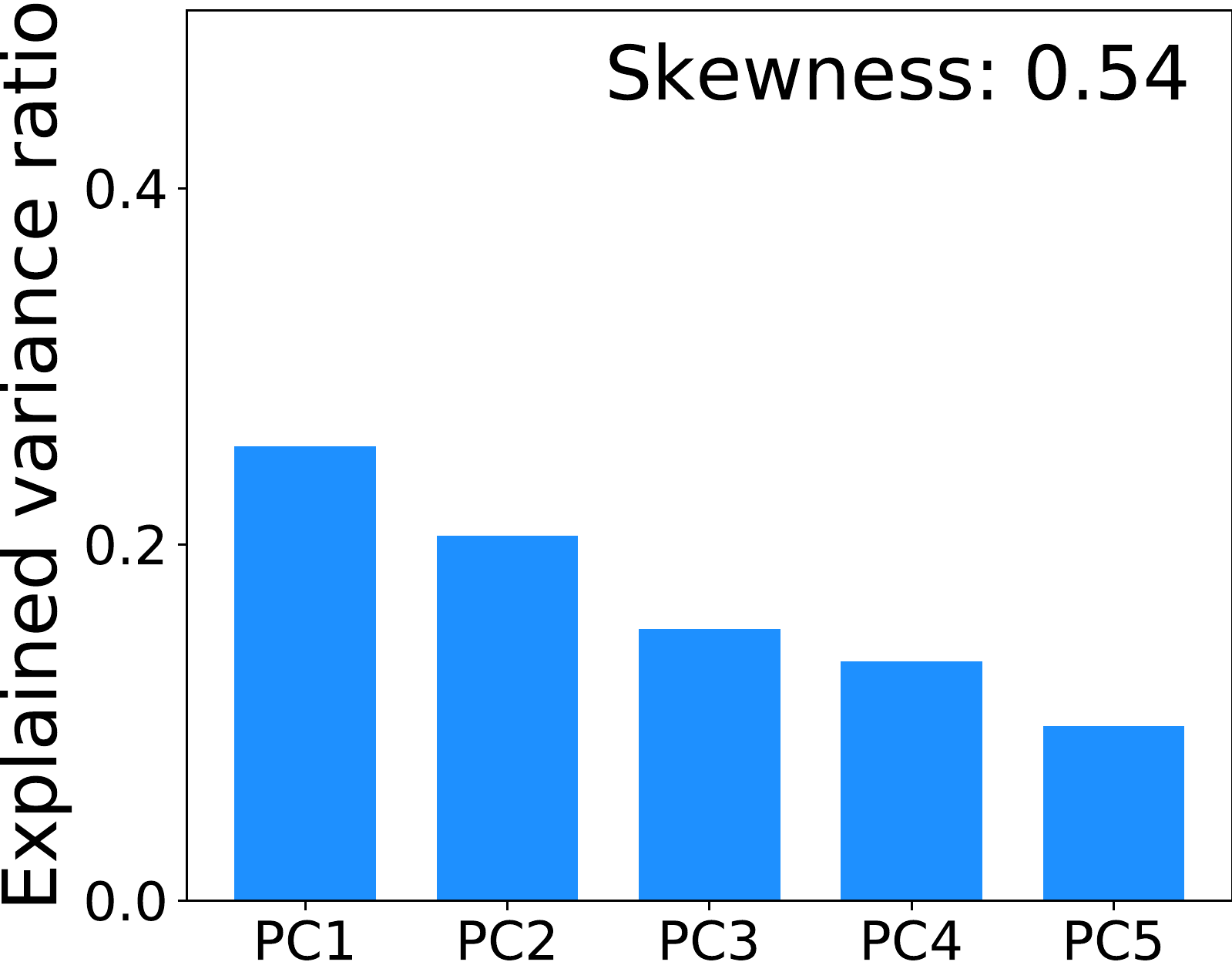}\label{fig:biasremoval:b}}
\hfill\null
\caption{\textbf{Bias removal in the feature space effect} on the explained variance of the principal components of $\bm{\Delta}$ on CIFAR-10S. After the removal of the bias direction, there is no longer a main direction of variance as illustrated by a reduced skewness.
}
\label{fig:biasremoval}
\vspace{-0.5em}
\end{figure}

\paragraph{Bias removal.}
Once the proposed model has been trained, we compute $\bm{\Delta}$ in Eq.~\ref{eq:delta} from the training set of CIFAR-10S. When performing a principal component analysis on $\bm{\Delta}$, we observe that there remains a main direction explaining the variance (Figure~\ref{fig:biasremoval:a}). Though, compared with the baseline model (Figure~\ref{fig:bias:a}), our model with protected embeddings reduces the skewness from 2.63 to 1.87. This effect is even more noticeable after the bias removal (Figure~\ref{fig:biasremoval:b}). Indeed, the skewness drops to 0.54 and there is no longer a main direction of variance.
The bias removal operation reduces the presence of the bias in the feature space.

\begin{figure}[h]
\centering
\begin{minipage}{.49\textwidth}
\hfill
\begin{overpic}[width=.9\linewidth]{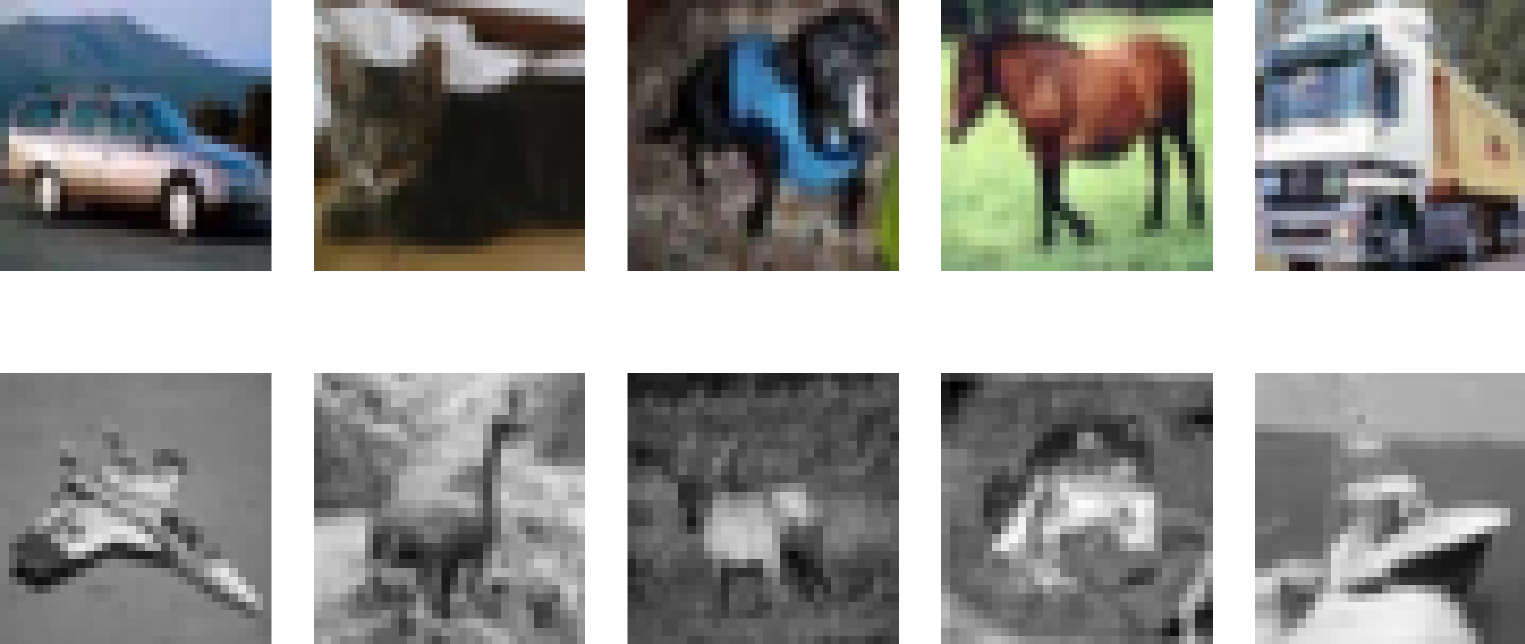}
\put(-8,28){\footnotesize\rotatebox{90}{color}}
\put(-8,5){\footnotesize\rotatebox{90}{gray}}

\put(-4,29){\footnotesize\rotatebox{90}{bias}}
\put(-4,5){\footnotesize\rotatebox{90}{bias}}

\put(5,20){\footnotesize\textit{auto}}
\put(27,20){\footnotesize\textit{cat}}
\put(47,20){\footnotesize\textit{dog}}
\put(66,20){\footnotesize\textit{horse}}
\put(86,20){\footnotesize\textit{truck}}

\put(1,-5){\footnotesize\textit{airplane}}
\put(27,-5){\footnotesize\textit{bird}}
\put(46,-5){\footnotesize\textit{deer}}
\put(67,-5){\footnotesize\textit{frog}}
\put(87,-5){\footnotesize\textit{ship}}
\end{overpic}
\vspace{.5em}
\caption{\textbf{CIFAR-10S samples}, where five classes are skewed towards color images and five other classes are skewed towards gray images in the training set.
\newline
}
\label{fig:cifar:samples}
\end{minipage}\hfill
\begin{minipage}{.49\textwidth}
\centering
\begin{overpic}[width=0.9\linewidth]{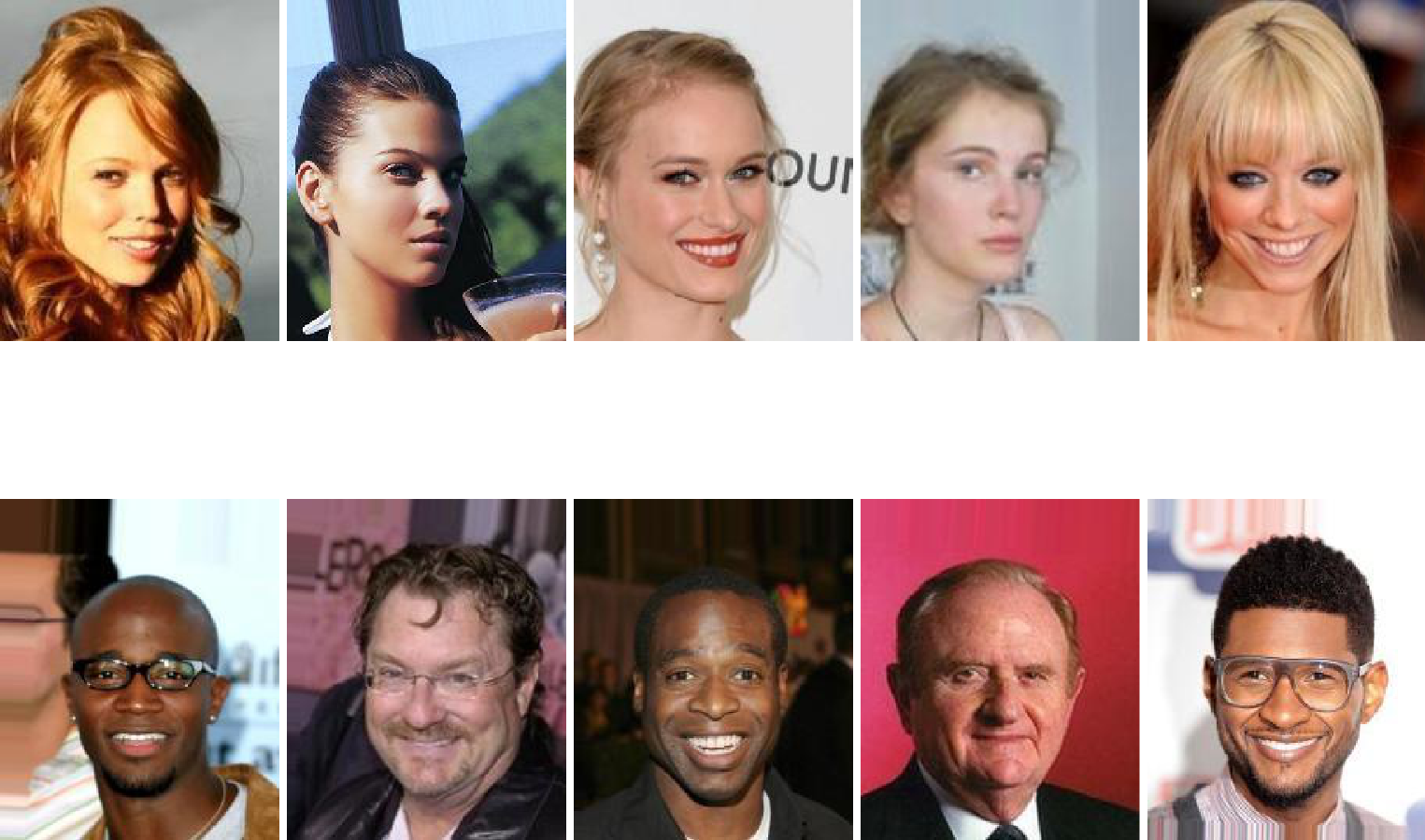}
\put(-8,40){\footnotesize\rotatebox{90}{female}}
\put(-8,7){\footnotesize\rotatebox{90}{male}}

\put(-4,42){\footnotesize\rotatebox{90}{bias}}
\put(-4,8){\footnotesize\rotatebox{90}{bias}}

\put(3,31){\footnotesize\textit{makeup}}
\put(24,31){\footnotesize\textit{lipstick}}
\put(40,31){\footnotesize\textit{rosy cheeks}}
\put(63,31){\footnotesize\textit{earrings}}
\put(81,31){\footnotesize\textit{blonde hair}}

\put(4,27){\footnotesize (99.7\%)}
\put(23,27){\footnotesize(99.4\%)}
\put(43,27){\footnotesize(97.9\%)}
\put(64,27){\footnotesize(96.5\%)}
\put(85,27){\footnotesize(94.3\%)}

\put(6,-4){\footnotesize\textit{bald}}
\put(20,-4){\footnotesize\textit{double chin}}
\put(44,-4){\footnotesize\textit{chubby}}
\put(62,-4){\footnotesize\textit{gray hair}}
\put(82,-4){\footnotesize\textit{eyeglasses}}

\put(4,-8){\footnotesize (99.8\%)}
\put(23,-8){\footnotesize(88.3\%)}
\put(43,-8){\footnotesize(88.1\%)}
\put(64,-8){\footnotesize(86.3\%)}
\put(85,-8){\footnotesize(79.5\%)}

\end{overpic}
\vspace{0.5em}
\caption{\textbf{CelebA samples} of the top-5 attributes skewed towards ``female'' and ``male'' genders in the training set.
}
\label{fig:celeba:samples}
\end{minipage}
\end{figure}

\paragraph{Samples from datasets.} Figure~\ref{fig:cifar:samples} shows examples of the color bias in CIFAR-10S~\cite{wang2020towards} while Figure~\ref{fig:celeba:samples} shows examples of the gender bias in CelebA~\cite{liu2015faceattributes}. 

\end{document}